\pdfoutput=1

\documentclass[11pt]{article}

\usepackage[]{acl}

\usepackage{times}
\usepackage{latexsym}

\usepackage[T1]{fontenc}

\usepackage[utf8]{inputenc}
\usepackage{graphicx}
\usepackage{subfig}
\usepackage{microtype}

\newcommand{\commentSH}[1]{\textcolor{black}{#1}}
\newcommand{\commentRZ}[1]{\textcolor{black}{#1}}
\newcommand{\reviseaacl}[1]{\textcolor{black}{#1}}
\title{Text Revealer: Private Text Reconstruction via Model Inversion Attacks against Transformers}

\author{Ruisi Zhang \\
University of California San Diego
  \And
  Seira Hidano \\
  KDDI Research, Inc.
  \And
  Farinaz Koushanfar \\
  University of California San Diego
}

\begin{document}
\maketitle
\begin{abstract}
Text classification has become widely used in various natural language processing applications like sentiment analysis. Current applications often use large transformer-based language models to classify input texts. However, there is a lack of systematic study on how much private information can be inverted when \commentRZ{ publishing models}. \commentSH{In this paper, we formulate \emph{Text Revealer} --- the first model inversion attack for text reconstruction against text classification with transformers. Our attacks faithfully reconstruct private texts included in training data with access to the target model. We leverage an external dataset and GPT-2 to generate the target domain-like fluent text, and then perturb its hidden state optimally with the feedback from the target model. Our extensive experiments demonstrate that our attacks are effective for datasets with different text lengths and can reconstruct private texts with accuracy.}

\end{abstract}

\section{Introduction} 
\label{sec:intro}

Natural language processing with its application in various fields have attracted much attention in recent years. 
With the recent advance in transformer-based language models (LMs), BERT~\cite{devlin2018bert} and its variants~\cite{liu2019roberta,xie2021elbert,qin2022bibert} are used to classify text datasets and achieve state-of-the-art performance. However, LMs tend to memorize data during training, which results in unintentional information leakage~\cite{carlini2021extracting}. 

Model inversion attacks~\cite{fredrikson2015model}, which invert training samples from the private dataset, has long been applied in the vision domain~\cite{yang2019adversarial, zhang2018collaborative, wang2021variational, kahla2022label}. For text-based datasets, the model inversion attacks have been applied in the medical domain to infer patients' privacy information. Recent work like KART~\cite{nakamura2020kart} and Lehman \textit{et al.}~\cite{lehman2021does} \reviseaacl{consider} inverting tabular data with sensitive attributes from the medical datasets. However, they follow a fill-in-blank scheme and fail to reconstruct sentences with fluency from scratch. 

\commentRZ{In this paper, we focus on \reviseaacl{reconstructing private training data from fine-tuned LMs at inference time.} It has a more general scenario where unauthorized personal data such as chats, comments, reviews, and search history may be used to train LMs. We perform a systemic study on how much private information is leaked via model inversion attacks.} There are several challenges when performing model inversion attacks on NLP models. First, the candidate pixel range is 256 in images, but the candidate token range is more than 30,000. Therefore, it is harder to find the exact token for the sentence. Secondly, image inversion is more error-tolerant, i.e., error in some pixels will not affect the overall results. However, errors in some of the tokens will significantly affect the fluency of the texts. Thirdly, current text reconstruction attacks have different settings from model inversion attacks. Most of them fall into two categories: Gradient attack~\cite{deng2021tag, zhu2019deep}, which utilize gradient during distributed training to do attack. Embedding level reconstruct attack~\cite{xie2021reconstruction, pan2020privacy}, which trains a mapping function to \reviseaacl{reconstruct texts from pre-trained embeddings}.



We propose \emph{Text Revealer} to \reviseaacl{perform model inversion attack on text data}. 
\commentRZ{In the attack, the adversary knows the domain of the private dataset and has access to the target models. Our attack consists of two stages: in the first stage, we collect texts from the same domain as the public dataset and extract high-frequency phrases from the public dataset as templates. Then, we train a GPT-2 as the text generator on the public dataset. In the second stage, we borrow the idea from PPLM~\cite{dathathri2019plug} to continuously perturb the hidden state in the GPT-2 based on the feedback from the target model. By minimizing the cross-entropy loss, generated text distribution becomes closer to the private datasets. Experiments on the Emotion and Yelp datasets with two target models, BERT and TinyBERT, demonstrate \emph{Text Revealer} can \reviseaacl{reconstruct} private information with readable contents.} In summary, our approach has the following contributions:

\noindent $\bullet$ We propose \emph{Text Revealer}, the first model inversion attack for text reconstruction against text classification with transformers, to \reviseaacl{reconstruct} private training data from the target models.

\noindent $\bullet$ Results on two transformer-based models and two datasets with different lengths have demonstrated \emph{Text Revealer} can \reviseaacl{reconstruct} private texts with accuracy.

\section{Approach} 
\label{sec:method}

\subsection{Threat Model}

\paragraph{Adversary's Target and Goal} 
The adversary's goal is to invert memorized training data from the fine-tuned LMs. \commentRZ{The inverted texts should meet two requirements: (1) the texts are close to the distribution of private dataset $D_{\mathrm{pri}}$. (2) the texts are readable so the adversary can infer meaningful contents from them.}  In our paper, we use the BERT and TinyBERT as the target LMs to do text classification tasks. We choose the target LMs for the following reasons: (1) BERT and its variants achieve state-of-the-art in most text classification tasks. (2) From the ethical standpoint, evaluating how sensitive the transformer-based model is to text classification, one of the essential tasks in the NLP domain, is important for subsequent research on defense methods.  

\paragraph{Adversary's Capability and Knowledge}
We consider the adversary knows the domain of the dataset on which the language model (LM) is fine-tuned. The adversary also has white-box access to the LM. During the attack, given the input sentences or input embeddings, the adversary can get the prediction score over $N$ classes with the probabilities $P = (p_1, p_2, ..., p_n)$. \commentSH{Throughout this paper, let $p_a(X)$ denote the prediction with a given input $X$ for a label $a$.}

\subsection{Attack Construction}

\begin{figure}
    \centering
    \includegraphics[width=\columnwidth]{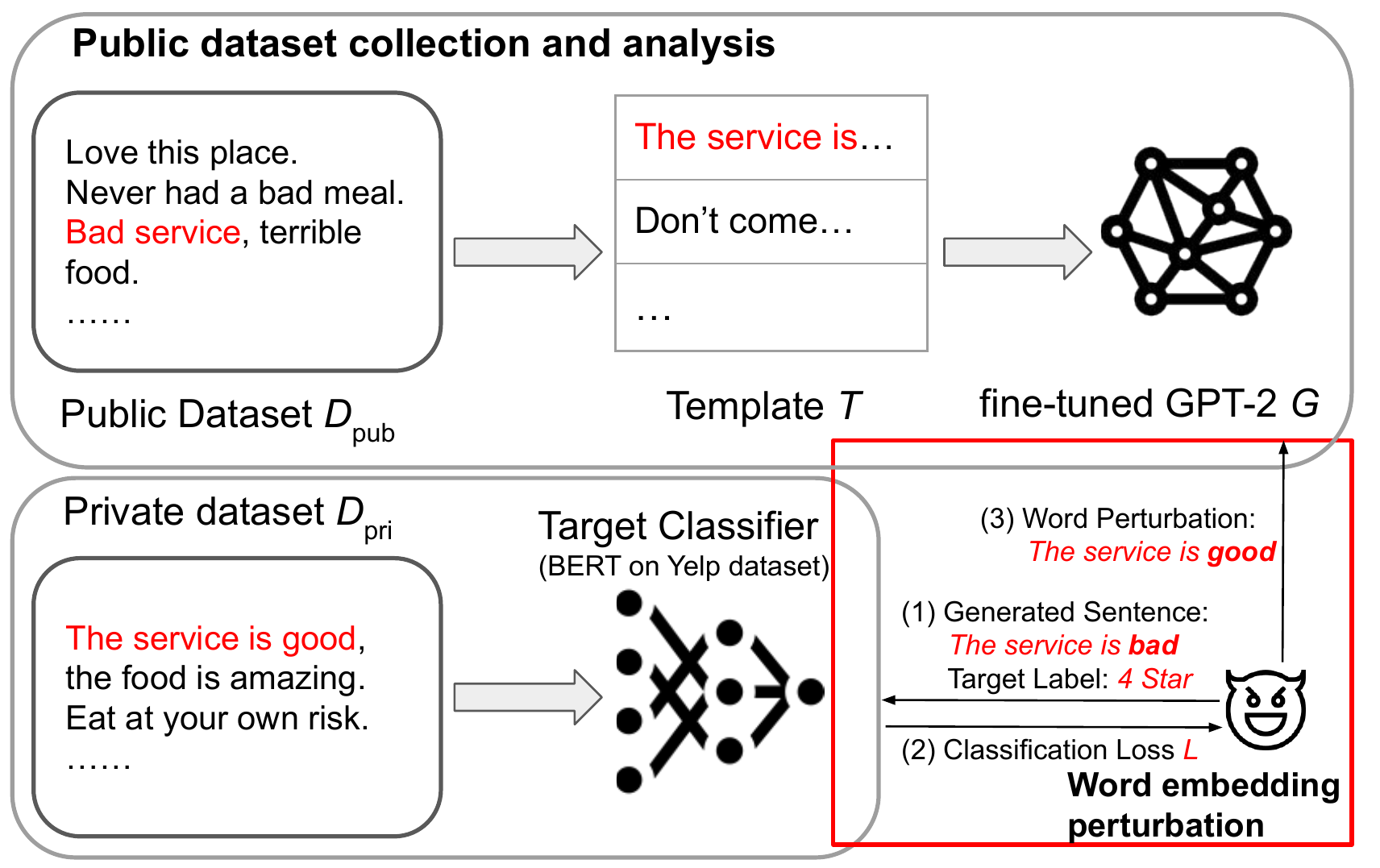}
    \caption{Attack construction pipeline.}
    \label{fig:pipeline}
\end{figure}

\commentRZ{As shown in Figure~\ref{fig:pipeline}, the general attack construction consists of two stages: (1) public dataset collection and analysis, and (2) word \reviseaacl{embedding} perturbation. }


\paragraph{Public dataset collection and analysis} 
\commentRZ{We first collect dataset from the same domain to form public dataset $D_{\mathrm{pub}}$ and perform $n$-gram analysis. There is no annotation in $D_{\mathrm{pub}}$. As the dataset grows larger, the word frequency between private dataset $D_{\mathrm{pri}}$ and public dataset $D_{\mathrm{pub}}$ becomes closer (see Section~\ref{a:analysis} for the details). Therefore, we can infer some \reviseaacl{phrases} may appear in the private dataset by doing the $n$-gram analysis and collecting high-frequency items as template $T$. Then, we fine-tune a GPT-2 $G$ to generate fluent texts following the distribution of $D_{\mathrm{pub}}$. We train $G$ by minimizing $\min_{G} L(G, D_{\mathrm{pub}})$, where $L$ is the cross entropy loss.}


\paragraph{Word \reviseaacl{embedding} perturbation}
We borrow the text perturbation idea from \reviseaacl{Plug and Play Language Model(PPLM)}~\cite{dathathri2019plug} but change the optimization objective to perform the model inversion attack. PPLM is a lightweight text generation algorithm that uses an attribute classifier to help the GPT-2 perturb its hidden state and guide the generation. 

\commentSH{We perturb the hidden state of the text generator to make generated texts closer to the distribution of the private dataset. Let $H_t$ be the current hidden state of the text generator $G$. Let $L_{\mathrm{adv}}$ be an adversarial loss to measure the distance between the generated text $G(H_t)$ and the private dataset $D_{\mathrm{pri}, a}$ of the target label $a$. The adversary generates perturbations $\Delta H_t$ for $H_t$ by solving the following optimization problem:}
\begin{equation}
\label{eqn:general}
\commentSH{\min_{\Delta H_t} L_{\mathrm{adv}}(G(H_t + \Delta H_t), D_{\mathrm{pri}, a}).}
\end{equation}

\commentSH{Since the adversary has no prior knowledge of the private dataset, we measure the loss $L_{\mathrm{adv}}$ with the cross-entropy calculated from the target model's prediction $p_a(G(H_t))$ with the generated text $G(H_t)$ for the target label $a$. This is because the cross-entropy takes a small value when the input to the model is similar to the training data. We also show that the cross-entropy is more effective for our model inversion attacks than the loss based on the whole prediction score $P$, such as the modified entropy~\cite{song2021systematic} (see Section~\ref{sec:ablation_studies} for the details). To obtain the optimal $\Delta H_t$, we minimize the cross-entropy by decending the hidden state.}

\section{Experiments} 
\label{sec:experiment}
\subsection{Datasets}
\textbf{Emotion Dataset}~\cite{saravia-etal-2018-carer} is a sentence-level emotion classification dataset with six labels: sadness, joy, love, anger, fear, and surprise. 
\textbf{Yelp Dataset}~\cite{zhang2015character} is a document-level review dataset. The reviews are labeled from 1 to 5 stars indicating the user's preference. Following the split methods in image model-inversion attacks~\cite{Zhang_2020_CVPR}, we randomly sample 80\% of the samples as public dataset and 20\% of the samples as the private dataset.  The average token length is 20 in the private Emotion dataset and 134 in the private Yelp dataset. 


\subsection{Evaluation Metrics}

\textbf{Recovery Rate (RR.)} is the percentage of tokens in private dataset recovered by different attack methods. We filtered punctuations, special tokens and NLTK's  stop words ~\cite{bird2009natural} in the private dataset. 
\textbf{Attack Accuracy (Acc.)} is the classification accuracy using evaluation classifier on inverted texts. According to ~\cite{Zhang_2020_CVPR}, the higher the classification accuracy is, the more private information the texts is considered to be inverted. We use BERT Large as the evaluation classifier and  fine-tune it until the accuracy on the private dataset is over 95\%.  
\textbf{Fluency} is the Pseudo Log-Likelihood (PLL) of fixed-length models on inverted texts\footnote{https://huggingface.co/docs/transformers/perplexity}.

\subsection{Baselines}
We compare our method with vanilla model inversion attack~\cite{fredrikson2015model} and  vanilla text generation~\cite{radford2019language}. 

\paragraph{Vanilla model-inversion attack (VMI)} In this setting, the adversary exploits the classification loss by adjusting text embeddings and returning the texts minimizing the cross-entropy loss. We set the text length to the average length of the private dataset, adjust the text embeddings for 50 epochs and run the same times as the number of templates. 

\paragraph{Vanilla text generation (VTG)} In this setting, the adversary uses a GPT-2 model to generate the texts. The GPT-2 is trained on the public dataset and conditioned on our collected templates. We limit the maximum length to the average length of the private dataset. \commentRZ{For the attack accuracy on VTG, we calculate the frequency of collected templates under different labels in the public dataset and use the label with the highest frequency as the target label. During the attack, only VTG requires extra annotations to benchmark its performance. For target model TinyBERT and BERT, we run VTG twice and use the results for two target models.}

\subsection{Target Models}
We use two representative transformer-based LM as our target model: (1) Tiny-BERT~\cite{bhargava2021generalization} and (2) BERT~\cite{devlin2018bert}. The TinyBERT has four layers, 312 hidden units, a feed-forward filter size of 1200, and 6 attention heads. It has 110M parameters. The BERT has 12 layers, 768 hidden units,  a feed-forward filter size of 3072, and 12 attention heads. It has 4M parameters.

\begin{table*}[h]
\small
\centering
\begin{tabular}{lccc|ccc|ccc|ccc}
& \multicolumn{6}{c|}{Emotion Dataset} & \multicolumn{6}{c}{Yelp Dataset}  \\\cline{2-13}
& \multicolumn{3}{c|}{Tiny-BERT} & \multicolumn{3}{c|}{BERT} & \multicolumn{3}{c|}{Tiny-BERT} & \multicolumn{3}{c}{BERT} \\\cline{2-13}
 & VMI & VTG & TR  & VMI & VTG & TR & VMI & VTG & TR &  VMI & VTG & TR\\
\hline
RR.(\%)  &27.54 & 61.23 & \textbf{67.36} & 34.33 & 59.42 & \textbf{67.76} &50.89 & 74.47 & \textbf{83.22} &75.56 &73.22 & \textbf{84.29}\\
Acc. (\%)         &72.04 & 74.50 & \textbf{84.14} & 73.90 & 73.65 & \textbf{85.54} &22.56 & 23.34 & \textbf{33.98} & 21.29& 22.16& \textbf{34.22}\\
PLL           &1493 & \textbf{43.31} & 80.79 & 2628 & \textbf{42.27} & 94.53 & 2391.09 & \textbf{16.82} & 62.75 & 4010.46 & \textbf{16.64} & 72.78\\
\end{tabular}
\caption{Results of the model-inversion attack, TR is short for our \emph{Text Revealer}.}
\label{tab:result}
\end{table*}

\subsection{Results}
The result of our model inversion attack is summarized in Table~\ref{tab:result}. We can make the following observations: (1) VTG and \emph{Text Revealer} tend to invert more tokens and achieve lower PLL compared with VMI. This is because VMI trains the text embeddings from scratch with random initialization, which results in not meaningful combinations of tokens. Even though many tokens can be recovered using VMI, private information still cannot be inferred from the private dataset. (2) Compared with VTG, our algorithm achieves higher recovery rates and higher attack accuracy. By perturbing the hidden state of trained GPT-2, our algorithm can infer more private information from the target model. (3) For smaller Emotion dataset, all three methods achieve high attack accuracy and can invert private information from the private dataset. \commentRZ{However, in the larger Yelp dataset, VMI and VTG's attack accuracy becomes near 20\%. It is close to random classification because only 5 classes are in the dataset.} (4) For both VMI and \emph{Text Revealer}, BERT's recovery rate and attack accuracy are higher than TinyBERT. It means more private information is memorized as the transformer becomes larger. 

\subsection{Ablation Studies}
\label{sec:ablation_studies}
\paragraph{Effectiveness of fine-tuning GPT-2}
In this setting, we fine-tune the GPT-2 on the public dataset and compare the results with vanilla GPT-2 on the Yelp dataset. The results are summarized in Table~\ref{tab:public}. The table shows that fine-tuned GPT-2 achieves higher recovery rate and attack accuracy. It means more sensitive information is revealed from fine-tuning.

\begin{table}[h]
\small
    \centering
    \begin{tabular}{c|c|c|c}
         & RR. & Acc. & PLL\\
       \hline
       fine-tuned GPT & 84.29 &34.22 & 72.78\\
       vanilla GPT &79.17 & 31.29 & 72.21 \\
    \end{tabular}
    \caption{Model inversion with guidance from fine-tuned GPT-2 and vanilla GPT-2 on BERT.}
    \label{tab:public}
\end{table}

\paragraph{Effectiveness of model inversion attack}
We analysis the effectiveness of word perturbation and loss function by comparing with different methods. For word perturbation, we compare \emph{Text Revealer}'s word perturbation with Gumbel softmax~\cite{jang2016categorical}. For Gumbel softmax, we first use fine-tuned GPT-2 to generate original sentences, and set coefficient for tokens in vocabulary. Then, we update the coefficient based on the cross-entropy loss using Gumbel softmax and update the input sentences. From the table, we can find Gumbel softmax and modified perturbation achieve similar attack accuracy. However, the recovery rate and fluency is lower than modified perturbation. 

For loss function, we compare cross entropy with Modified entropy loss ~\cite{song2021systematic}. The modified entropy loss makes the loss monotonically decreasing with the prediction probability of the correct label and increasing with the prediction probability of any incorrect label. In this setting, we use the same pipeline as \emph{Text Revealer}, but change the loss to modified entropy loss to update GPT-2's hidden state. We can find cross entropy loss achieves best performance out of other loss functions.

\begin{table}[h]
\small
    \centering
    \begin{tabular}{c|c|c|c}
        Attack & RR. & Acc. & PLL\\
       \hline
       \emph{Text Revealer} & 84.29 &34.22 &72.78\\
       Gumbel softmax & 78.13 & 33.49 & 77.70\\
       Modified Entropy Loss & 74.23 & 32.74 & 77.25\\
    \end{tabular}
    \caption{Model inversion with different method.}
    \label{tab:attack}
\end{table}

\subsection{Analysis}
\label{a:analysis}

\paragraph{Effectiveness of template length}
In this setting, we analyze how the inversion accuracy changes with the length of the template and summarize the results in Table~\ref{tab:template}. The template length is $L$, and we extract the first 0.3 $L$, 0.5 $L$, and 0.7 $L$ tokens from the template. If the extracted token number is smaller than 1, we choose the first token from the template. For example, if the original template is "if i could give this place 0 star", then 0.7 $L$ template would be "if i could give this," 0.5 $L$ template would be "if i could give," and 0.3 $L$ template would be "if i."
As text length becomes shorter, \emph{Text Revealer} achieves worse recovery rates and attack accuracy. It means longer template length can give GPT-2 more contexts and help it invert more private information. 

\begin{table}[h]
\small
    \centering
    \begin{tabular}{c|c|c|c}
        Template Length & RR. & Acc. & PLL\\
       \hline
       0.3 $L$ & 84.17 & 17.60 & 76.45\\
       0.5 $L$ & 84.03 & 22.49 & 80.85 \\
       0.7 $L$ & 83.62 & 27.32 & 79.71 \\
        $L$ & 84.29 &34.22 &72.78\\
    \end{tabular}
    \caption{Model inversion with different length of the template. The percentage means the number of tokens we take since the start token.}
    \label{tab:template}
\end{table}

\begin{table*}[h]
    \centering
    \begin{tabular}{c|p{6cm}|p{6cm}}
         &  Ground Truth & Inverted \\
         \hline
       Example 1  & i don t \textbf{feel} so \textbf{exhausted} \textbf{all the time} &  \textbf{feel} a little \textbf{exhausted} feel a little bit from \textbf{all the time}\\\hline
       Example 2  & \textbf{i feel blessed amazed and} yes very excited & \textbf{i feel blessed and} grateful for the \textbf{amazing} new life \\\hline
       Example 3 & \textbf{the Parlor has} one of \textbf{the best} atmospheres \textbf{in Phoenix}! & \textbf{The Parlor has} soo, \textbf{the best} Mexican restaurants \textbf{in Phoenix} Palace! \\\hline
       Example 4 & The decor was very inviting and \textbf{we love Mexican food} (again, \textbf{from San Diego}) \textbf{The chips and salsa were} delicious.& \textbf{We are from San Diego} I \textbf{was a Mexican food lover}, we had \textbf{the chips, salsa was} very nice. \\
    \end{tabular}
    \caption{Iverted Examples, the first two are from the Emotion dataset, and the last two are from the Yelp dataset.}
    \label{tab:inverted}
\end{table*}

\begin{figure*}[h]
    \centering
    \subfloat[Example 1]{\includegraphics[width=0.25\linewidth]{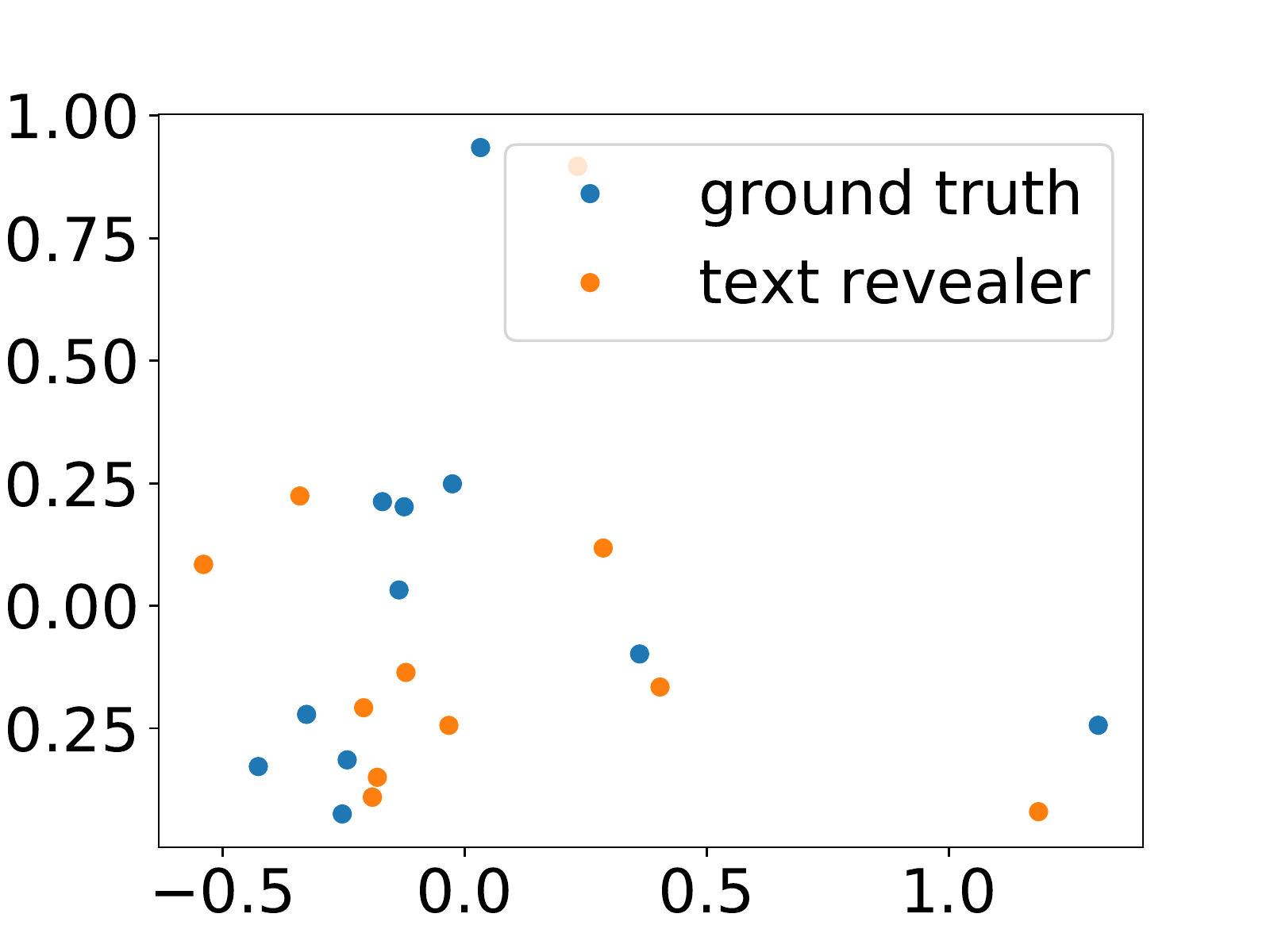}}
    \subfloat[Example 2]{\includegraphics[width=0.25\linewidth]{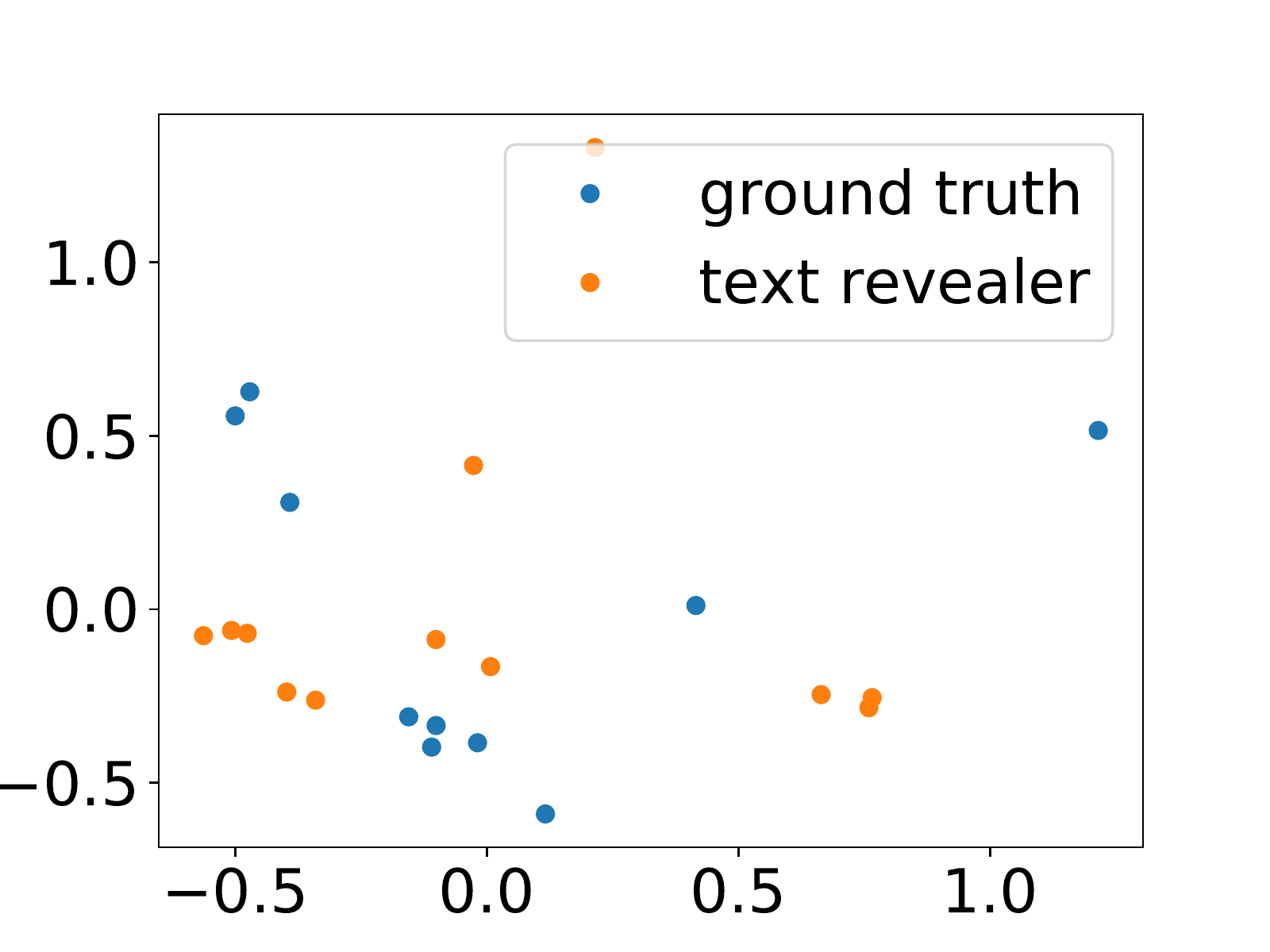}}
    \subfloat[Example 3]{\includegraphics[width=0.25\linewidth]{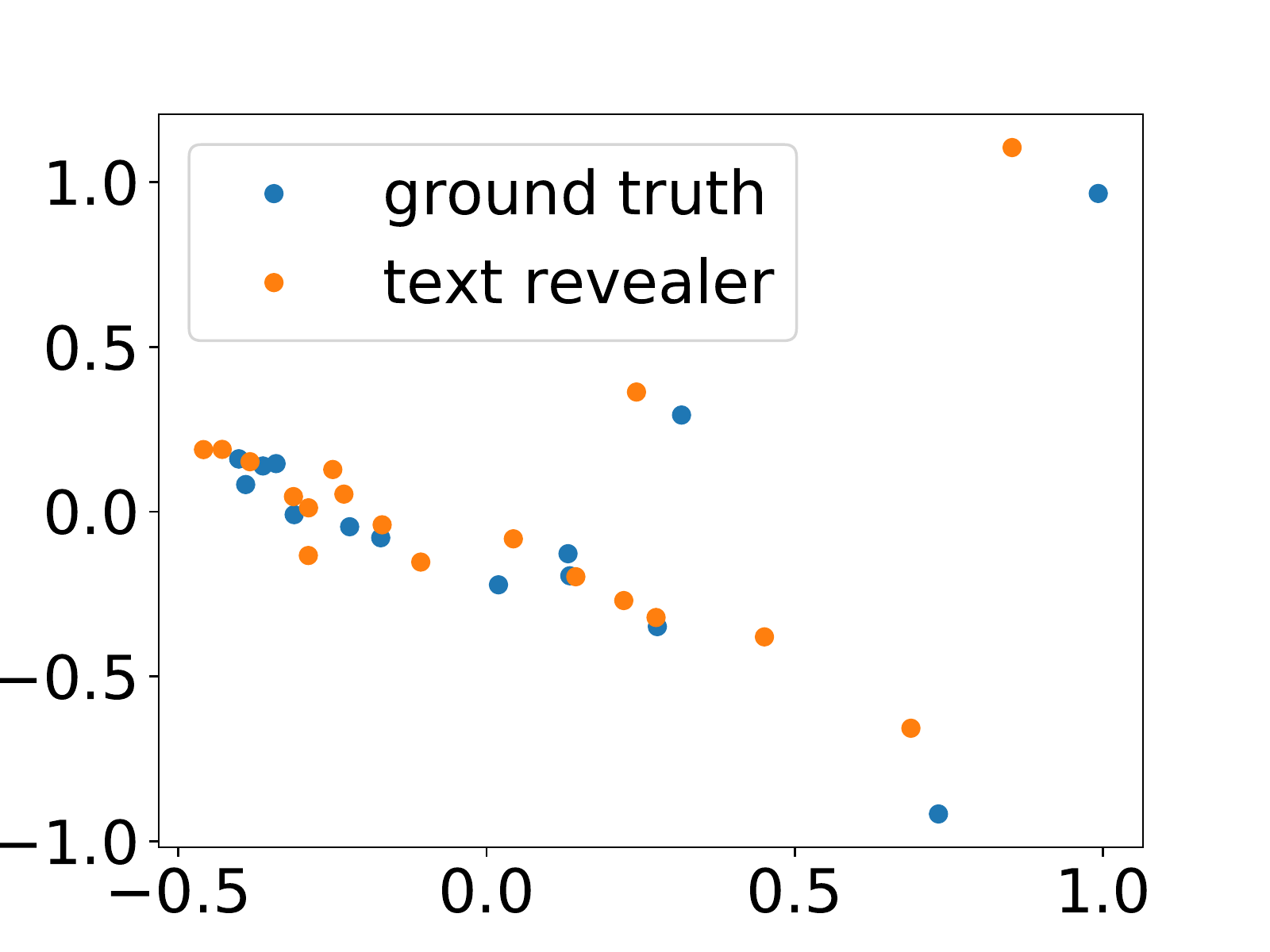}}
    \subfloat[Example 4]{\includegraphics[width=0.25\linewidth]{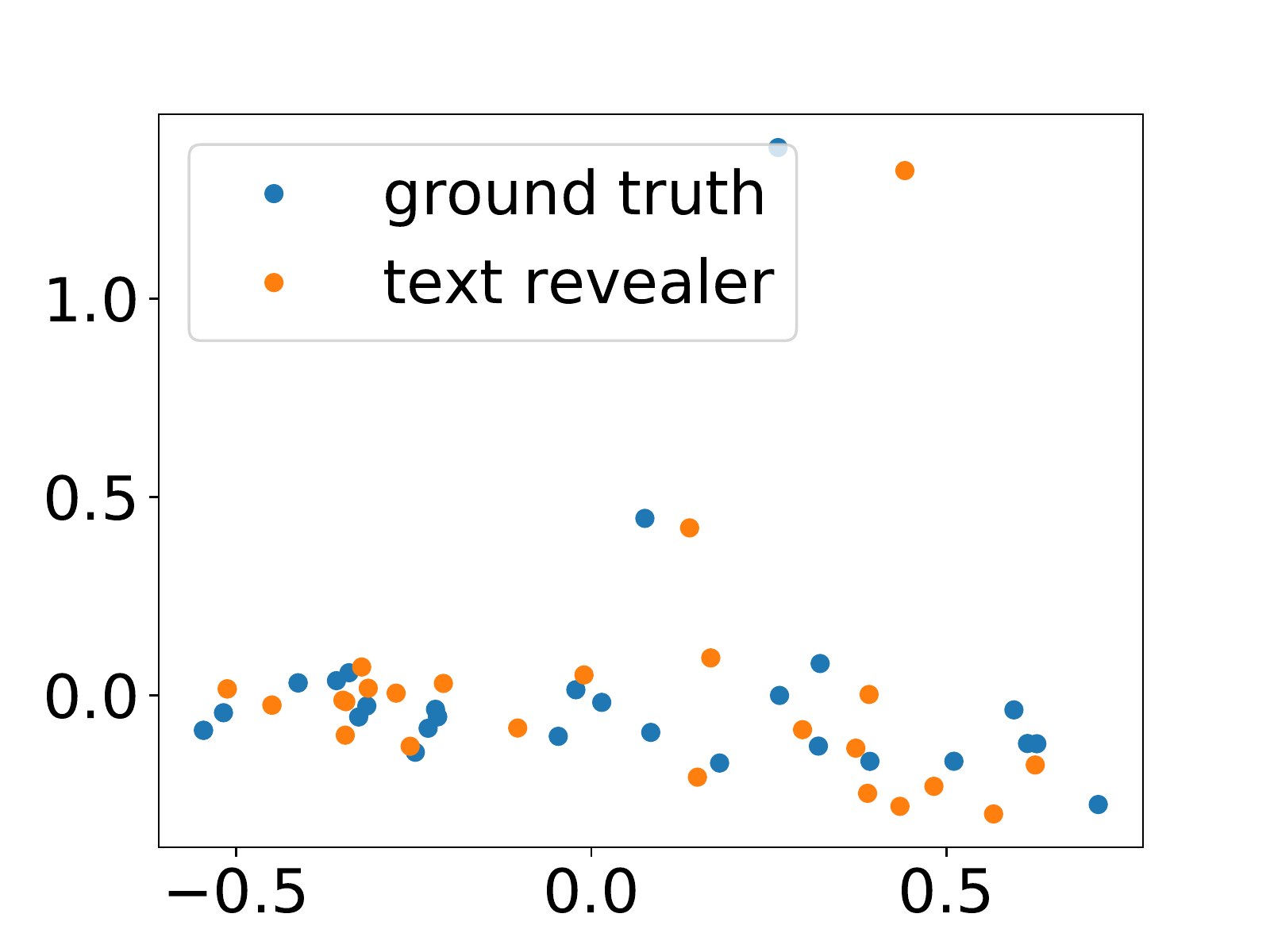}}
    \caption{Visualization of inverted embeddings, the first two are from the Emotion dataset, and the last two are from the Yelp dataset.}
    \label{fig:inverted}
\end{figure*}

\paragraph{Inverted examples \& Visualization}
Following TAG~\cite{deng2021tag}, We display how our inverted examples is approximate the distribution of private dataset at embedding level and sentence level. We use sentences with longest matching subsequence in the private dataset as ground truth. We first use Principal Component Analysis (PCA)~\cite{abdi2010principal} to reduce the dimension for ground truth and inverted examples and display the plot in Figure~\ref{fig:inverted}. Then, we display the inverted examples at sentence level in Table~\ref{tab:inverted}

\paragraph{Public dataset \& Private dataset}


We use Kendall tau distance~\cite{kendaldiatance} to measure the token distribution similarity between the private and public datasets. The Kendall tau distance is a metric to measure the top-$k$ elements correlation between two lists. The larger the distance is, the stronger correlation the two lists have. When ranking the tokens in each dataset, we filtered the special tokens, punctuations, and stop words in NLTK~\cite{bird2009natural}. \commentRZ{ From the results in Table~\ref{tab:sample}, we can find the word ranking and frequency is close between the private and public datasets.}

We also compare the accuracy of the Emotion and the Yelp Dataset in Table~\ref{tab:sample}. Before fine-tuning, the classification accuracy is similar for both private and public datasets. However, after fine-tuning, the accuracy between the private dataset and the public dataset has a large gap, which means some texts are memorized~\cite{carlini2021extracting} in the trained transformers.

\begin{table}[h]

    \centering
    \begin{tabular}{c|c|c}
     & Emotion & Yelp Dataset  \\ 
     \hline
    Distance(Top10)& 0.99 & 0.99 \\ 
    Distance(Top100)& 0.32 & 0.62 \\ 
    \hline
    Tiny-BERT-Public(+)  & 84.79 & 60.58\\ 
    Tiny-BERT-Private(+) & 94.15 &  79.44 \\ 
    \hline
    Tiny-BERT-Public(-)  & 12.81 & 20.38\\ 
    Tiny-BERT-Private(-) &13.25 & 20.59 \\ 
    \hline
    BERT-Public(+)  & 91.25  & 64.42 \\ 
    BERT-Private(+) & 99.66  & 95.12 \\ 
    \hline
    BERT-Public(-)  & 13.34 & 20.02 \\ 
    BERT-Private(-) & 14.09 & 19.88\\ 
    \end{tabular}
    \caption{The first cell is the ranking distance between the private and public datasets. The second and third cell is the accuracy on public and private datasets, "+" means the BERT model is fintuned on the private dataset, "-" means the plain BERT model.}
    \label{tab:sample}
\end{table}



\section{Conclusion} 
\label{sec:experiment}
In this paper, we present \emph{Text Revealer} to invert texts from the  private dataset. Experiments on different target models and datasets have demonstrated our algorithm can faithfully reconstruct private texts with accuracy. 
In the future, we are interested in exploring potential defense methods.

\bibliography{anthology,custom}
\bibliographystyle{acl_natbib}

\clearpage
\setcounter{page}{1}
\appendix
\section{Appendix}
\label{sec:appendix}
\subsection{Limitations}
Our work has the following limitations. (1) From the attack side, the model inversion attack achieves lower attack accuracy as the dataset becomes larger. It would be interesting to explore better template extract methods to improve the accuracy. (2) From the ethical standpoint, evaluating how sensitive BERT based model is to classification tasks is essential to future defense methods. Due to page limits, we did not go into the defense methods. We will continue to work on it in future work.

\subsection{Ethics Statement}
Our work focuses on the privacy problems in natural language processing. Even though model inversion attacks proposed in our paper can cause potential data leakage. Evaluating how sensitive BERT based model is to classification tasks is important to future defense methods.

\subsection{Hyperparameters and training details}
\label{a:hyper}

\paragraph{Template} We perform $n$-gram analysis on the public dataset. For Emotion dataset, we extract phrase length from 1 word to 3 words with frequency higher than 20. Then, we filter one-word adjectives and permute them with 2-3 word to form template.
The total number is 220. For Yelp dataset, we extract phrase length from 3 words to 8 words with frequency higher than 20. The total number is 320. Some of the templates are summarized in Table~\ref{tab:template}

\begin{table*}[htbp]
    \centering
    \begin{tabular}{c|p{15cm}}
      Datasets  & Examples \\
      \hline
      Emotion   & feeling a little, feel like, down, alone, love, good, passionate, sweet, irritable, angry, strange, nervous, surprised, amazed \\
      \hline
      Yelp      & if i could give this place zero stars, i will never go, really wanted to like this place, i cant wait to go back, if you are looking for, cant say enough good things about, one of the best, i m not a fan of, this place is, if you re in the mood for, only reason i didn t give it\\
    \end{tabular}
    \caption{Template examples.}
    \label{tab:template}
\end{table*}

\paragraph{BERT fine-tuning} The BERT and TinyBERT are trained as target models on the private dataset. For both BERT and TinyBERT, we set the batch size to 8. We use AdamW as the optimizer with initial learning rate set to 5e-5. We train 5 epochs on BERT and 10 epochs on TinyBERT. Other parameters follow the original implementation in ~\cite{devlin2018bert, jiao2019tinybert}.

\paragraph{GPT-2 fine-tuning} The GPT-2 is trained on the public dataset to help \emph{Text Revealer} reconstruct texts. For GPT-2, we set the batch size to 8, and use AdamW as the optimizer with initial learning rate
set to 5e-5. We train 10 epochs for both datasets. Other parameters follow the original implementation in ~\cite{radford2019language}. It has 124M parameters.

\paragraph{Text Revealer} When using the \emph{Text Revealer}, we set the iteration number to the average length of the private dataset. We set the window mask to 3 and KL loss coefficient to 0. Other parameters follow the original implementation in ~\cite{dathathri2019plug}.

\paragraph{Computing Infrastructure}

Our code is implemented with PyTorch. Our attack pipeline are all constructed using TITAN Xp. We fine-tune the target LMs and GPT-2 on NVIDIA RTX A6000.

\subsection{More inverted examples}
\label{a:example}

We display more inverted examples in Table~\ref{tab:examples} and Table~\ref{tab:examples2}. We show four examples for each method in Emotion dataset and two examples for each method in Yelp dataset.

\begin{table*}[htbp]
    \centering
    \begin{tabular}{c|p{15cm}}
      Method  & Examples \\
      \hline
      BERT-VMI  &  cancel the dream vascular place seeing halfway hopeful applicant Mandy. \newline Higgins suspiciously information poollifting Cardinal faint Scots Davidfeld \newline lanes creepy fears Call lap dick Evening Slim favor \newline impressive wireless baptism longing Independent IN forensic Wish physically\\
      \hline
      BERT-VTG  & \textbf{feeling a little} inspired by their own experience. \newline \textbf{feel like productive} people and more than a little creepy.\newline \textbf{i feel reluctant} to give me their phone number. \newline \textbf{i feel uncomfortable} at all in the light of it\\
      \hline
      BERT-TR &   \textbf{feel like glad} to be here i feel welcomed. \newline \textbf{feel like mellow} at breathlessly flowing smoothly and i feel was was pretty \newline \textbf{i feel afraid} that people dont see the movie spoiler place where \newline \textbf{feel uncomfortable} talking to my partner or to stop feeling guilty 
      \\
      \hline
      TinyBERT-VMI  & roughly spiritual thereof 41 rebellious fragrance its caused strangled ave  \newline algeria toward puzzled kowalski bobbie sexually liz circle hiding \newline code balls redundant persuaded carpathian yugoslav drownedpins \newline  burnley rubbish receipts borough fragmentation worthless soga\\
      \hline
      TinyBERT-VTG  & \textbf{i feel resentful} and lonely, she doesn't seem to like me \newline \textbf{feel like irritated} that my boss is complaining about something \newline \textbf{feeling a little} rude but i wont be afraid \newline \textbf{feeling a little pain} in my lower back and im trying to harden  \\
      \hline
      TinyBERT-TR & \textbf{feeling a little} amazing artistic life positive super job  productive \newline \textbf{feeling mellow} wise just like how much she is \newline \textbf{I feel thankful} that god gave me more blessings in return \newline \textbf{feel like overwhelmed} or at the wrong spot saying good feeling\\
    \end{tabular}
    \caption{Emotion dataset inverted examples, TR is short for our \emph{Text Revealer}.}
    \label{tab:examples}
\end{table*}

\begin{table*}[htbp]
    \centering
    \begin{tabular}{c|p{14cm}}
      Method  & Examples \\
      \hline
      BERT-VMI  & promising odor Daemon asked exactly platinum wonderful  confession Aaron Tiger Magazine field AMC documentaries edible Name Always Benito Northeast invalid Brooks d Games exist range removal thee drag monkey revolution hostile graphicspent permitted expenses mythical newest Massey\newline  \#\#wood successors rich hood Jameson Lin publishes festivities Slave ordering trans plot text Programs reluctantly promising Glad España stands combining  accommodate captain elongated stranger Conservatory express receives phones bias Depending pile motivation Loving chickens owl denied bage \\
      \hline
      BERT-VTG  & \textbf{i wouldn t go out of my way} to come here, but i had a coupon for a 50 dollar pizza for one hour and a free 30 minute massage in addition to the 20 dollars i spent.  when we arrived the staff greeted us and escorted us to our room.  that was good service and i enjoyed that.  \newline
      \textbf{can t wait to go back} and try more!  the food was amazing and the service was beyond fast!  I had the grilled pork chop which was one of the best dishes I have ever had in my life.  the other dishes, the veal, were also very good!  The dessert was just as amazing, I would have happily eaten any dessert.
      \\
      \hline
      BERT-TR &  \textbf{i wouldn t go out of my way} to come here, but i love the food and atmosphere well, i'm a foodie, so i know what to expect from this place. the chips are the BEST i've had in my lifetime. i had the beef barbacoa taco and i know what to expect from this place, it was absolutely delicious!\newline \textbf{i can t wait to go for} me to get my first! This is the most of my favorite place, you. I was shopping a place to take my family that and get the best. the restaurant is always clean and the staff are friendly.
      \\
      \hline
      TinyBERT-VMI  & \#\#wil apt fish pairs certification maker walker indictment thieves unity negligence 69 crook criminal fraud aligned partly negligence racist bewildered architects inspectors pageant fraud victims atheist restore  harassment gasoline ny removed screenplay transformed feathers investigation regular layers random \newline haiti sloping revolving simulator visitor decided marathon faerie sizable  sob ty gerald decker buzz shortage suriname verity dream edpace batavia oliver meter41 winced surround fewer grumbled celtics reducing  qualify secured completes brandon dimly brandonnb pointless hunts proper attemp\\
      \hline
      TinyBERT-VTG  &  \textbf{i can t say enough good things about} this     place i just  will be back    and i promise you i will never miss a thing     this place is the bomb. and the staff is super nice and helpful post again.the good is amazing. they have a wide variety of beers all great.even on tap. and the service is top notch \newline \textbf{if i could give this place zero stars}i would. i have been going here since it opened, but in recent years this place has went downhill. the waiters here use to be very attentive and nice. now i don't get it. everything they bring you seems to have a strange \"yippie\" taste, which makes my blood boil. 
      \\
      \hline
      TinyBERT-TR & \textbf{you get what you pay for} for! i will always give my money to a place that gives me the best quality I can give. the ambiance is amazing, but the food takes forever. one friend and i went on a tuesday night, and although the place was dead, only a handful of people were eating. order 2 things: the kobe beef shortribs, the shish kabob, and the chicken shawerma. \newline \textbf{can t wait to go back}! my husband had the shrimp tacos and they were fantastic! he had the margaritas and they were ok, but, the atmosphere and staff was just really nice and friendly. we'll definately go back. for the awesome margaritas!!for the delicious food! great prices!!!to our server, jason. for the amazing service! \\
    \end{tabular}
    \caption{Yelp dataset inverted examples, TR is short for our \emph{Text Revealer}.}
    \label{tab:examples2}
\end{table*}


\end{document}